\documentclass[10pt, a4paper]{article}

\usepackage[final]{lrec-coling2024} 

\usepackage{enumitem}

\title{Hands-On Tutorial: Labeling with LLM and Human-in-the-Loop}

\name{Ekaterina Artemova$^1$  \quad Akim Tsvigun$^2$  \quad Dominik Schlechtweg$^3$ \\
\large\bf {Natalia Fedorova$^1$  \quad Sergei Tilga$^1$ \quad Konstantin Chernyshev$^1$ \quad Boris Obmoroshev$^1$} }

\address{$^1$ Toloka AI, $^2$ Nebius AI,  $^3$University of Stuttgart \\
         \small{\textbf{Correspondence}: katya-art@toloka.ai}\\
        \small{\textbf{Web page}: \url{https://toloka.ai/coling-2025-human-w-llm-tutorial}}\\}

\abstract{
Training and deploying machine learning models relies on a large amount of human-annotated data. As human labeling becomes increasingly expensive and time-consuming, recent research has developed multiple strategies to speed up annotation and reduce costs and human workload: generating synthetic training data, active learning, and hybrid labeling. This tutorial is oriented toward practical applications: we will present the basics of each strategy, highlight their benefits and limitations, and discuss in detail real-life case studies. Additionally, we will walk through best practices for managing human annotators and controlling the quality of the final dataset. The tutorial includes a hands-on workshop, where attendees will be guided in implementing a hybrid annotation setup. This tutorial is designed for NLP practitioners from both research and industry backgrounds who are involved in or interested in optimizing data labeling projects.
 \\ \newline  }

\begin{document}

\maketitleabstract

\section{Introduction}

Labeled data is the primary tool for providing supervision to learnable systems. In the last decade, labeling data on crowdsourcing platforms or by hiring expert annotators has become the de facto standard approach to creating training data. However, human labeling can be time- and resource-consuming, hindering the ability to scale data labeling projects. Recent research has developed multiple strategies to use machine learning methods with human-in-the-loop to obtain larger amounts of high-quality data in shorter periods.

This tutorial explores recent strategies to speed up annotation and reduce costs and human workload. Emerging automatic approaches use language models (LMs) to \textit{generate synthetic training data}~(\S 3.2). By fine-tuning or prompting LMs, we can create data that closely mirrors the distribution of the target dataset. This synthetic dataset can be further used to train models from scratch or to augment existing datasets, particularly in low-resource settings. However, this approach may result in biased or low-quality data, highlighting the need for human efforts in labeling.

Automatic approaches for data labeling cannot replace entirely human annotation. Limitations of automatic approaches (\S 3.6) lie in the nature of the language models used and are associated with bias in generated data, the difficulty in accurately mirroring the nuances of real-world data, and the challenges in processing subjective, culture-specific, or knowledge-intensive domains.

Maintaining a human-in-the-loop for data labeling is essential to ensure high-quality and accurate annotations. By involving humans in the labeling process, we can continually adapt to new data, and address edge cases. \textit{Active learning} (\S 3.3) selects the most informative instances for human labeling to maximize model performance. \textit{Hybrid labeling} (\S 3.5) combines human and model efforts, with the model handling simple instances and humans handling complex ones. These approaches reduce the amount of labeled data needed, lowering the annotation budget and improving data accuracy.

The quality of the resulting dataset depends on automatic methods and effective management of human workers. We will cover best practices in writing guidelines and using quality control methods (\S 3.4). Additionally, we emphasize the importance of maintaining ethical standards when working with human annotators, ensuring fair payment, continuous communication, and avoiding conditions harmful to their well-being.

The tutorial includes a 30-minute hands-on session (\S 3.7), where the instructor will cover the setup of a hybrid data annotation project step-by-step. This session will feature a code walk-through and a real-life case study demonstrating how combining human and LM-based labeling yields high-quality data affordably and quickly.

Aimed at NLP practitioners involved in data labeling projects, this tutorial focuses on practical applications. Each section of the tutorial will feature a real-life case study to demonstrate labeling approaches. These case studies, drawn from the instructor's research or industrial experience, provide practical insights and share effective strategies.

By the end, participants will understand how to optimize their annotation workflows for natural language processing tasks, such as text or token classification.


\section{Outline}

\subsection*{Part 1: Introduction (20 min)}




\begin{itemize}[nosep]
  \item \textbf{Introduction}: discussing the need for large amounts of labeled data
  \item \textbf{Key concepts}: annotation schema and guidelines, quality control, inter-annotator agreement.
  \item \textbf{Annotation lifecycle:} the typical setup of an annotation project; transition from human workers to LLMs.
  \item \textbf{Costs and duration of an annotation project}: comparison of projects with and without using LLMs.
  \item \textbf{Benefits and limitations of LLMs}: reduced time and labor but lower quality.
  \item \textbf{Conclusion}: motivating the need for LLMs and human-in-the-loop approaches, overview of next sections.
\end{itemize}

\subsection*{Part 2: LM workflows (30 min)} \label{subsec:workflow}

\textbf{Overview} This section will demonstrate best practices for common workflows involving language models (LMs) and large language models (LLMs). These workflows aim to (i) create efficient LMs with acceptable performance optimized for labeling data, and (ii) generate synthetic data for data augmentation. 

\begin{itemize}[nosep]

\item \textbf{Models}:  large-scale LMs, such as FLAN \cite{longpre2023flan}, Llama 3.1 \cite{llama2024herd} or XLM-R \cite{conneau2020unsupervised}
\item \textbf{Performance optimization}: prompt engineering; parameter-efficient fine-tuning \cite{lialin2023scaling}.
\item \textbf{Efficiency optimization}: quantization \cite{tao2022compression}; pruning \cite{wang2020structured}; knowledge distillation \cite{gu2024minillm}.
\item \textbf{Synthetic train data generation}: attribute-aware generation \cite{yu2024large}, zero-shot generation \cite{ye2022zerogen},  model collapse caused by training on synthetic data only \cite{shumailov2023curse}.

\item \textbf{Case study}: Sentiment analysis in media monitoring context. 

\end{itemize}


\subsection*{Part 3: Active learning with LMs (40 min)}

\noindent \textbf{Overview} This section presents Active learning (AL) in data annotation. We discuss key strategies for both generative and non-generative AL,  their applications, advantages, and limitations.

\begin{itemize}[nosep]
\item \textbf{Introduction to AL}: definition and core principles of AL; benefits of AL in reducing annotation costs and improving model performance; AL workflow.

\item \textbf{AL strategies for text classification}: basic strategies (Least Confidence~\cite{lc_lewis}, Breaking Ties~\cite{bt_luo}, etc.); gradient-based strategy BADGE~\cite{badge}, contrastive AL~\cite{cal_margatina}.

\item \textbf{Generative AL with LLMs}: leveraging LLMs to generate informative samples; controlling LLM-generated samples' diversity and relevance.

\item \textbf{Human-based vs. LLM-based annotation in AL}: strengths and weaknesses; strategies for combining human and LLM annotations; quality control measures for AL pipelines.

\item \textbf{Efficient AL}: methods to accelerate the acquisition stage; batched-mode AL.

\item \textbf{Case study:} AL for text classification in the law domain.

\end{itemize}

\subsection*{Part 4: Quality control and managing human workers (30 min)}

\noindent \textbf{Overview} This section focuses on quality control and best practices in working with human annotators. 

\begin{itemize}[nosep]
  \item \textbf{Pipeline setup}: sourcing \& selection of supply of annotators (including smart matching for the general crowd based on skills and demographics); onboarding, exams \& training of annotators; control tasks \cite{sabou2014corpus,gordon2022jury}.
  \item \textbf{Data production}: dynamic overlaps; platform tools for experts (time keeping, copy-paste \& word count limitations, spell check, etc.); anti-fraud rules \cite{drutsa2020crowdsourcing}, regular audits.
  \item \textbf{Data acceptance and working with data}: automatic metrics; audit by annotation or domain experts; aggregation tools for general crowd tasks; ML assessment of dataset quality; LLM-checks \cite{klie2024analyzing}.
  \item \textbf{Managing human annotation work}: writing annotation guidelines; psychological characteristics of doing annotation tasks; inter-annotator agreement; communication with the workers. \cite{saal1980rating,tseng2020best}.
\end{itemize}

\subsection*{Part 5: Hybrid pipelines (40 min)}

\noindent \textbf{Overview} This section presents developing hybrid pipelines, e.g. effectively combining human and model labeling to achieve the best balance of quality, cost, and speed.

\begin{itemize}[nosep]


\item \textbf{Model confidence estimation}: prediction quality depending on various confidence levels; confidence estimation in open-source and proprietary models \cite{geng2023survey,mukherjee2020uncertainty}.

\item \textbf{Aggregation of model and human responses}: setting confidence thresholds to route tasks between human and model labeling; combining model and human responses using overlaps, human response confidence, weighting techniques \cite{wei2023aggregate}.

\item \textbf{Balancing between quality and automation}: strategies for balancing quality cost, and speed of labeling through confidence thresholding; optimal combinations to maximize cost-efficiency and quality for various setups.

\item \textbf{Hybrid pipelines}: quality assessment; continuous improvement of model performance \cite{wang2024human}.

\item \textbf{Case study}: hybrid labeling applications for product recommendations and ad search relevance.


\end{itemize}

 \label{subsec:human_with_llm}

\subsection*{Part 6: Limitations (20 min)}

\textbf{Overview}   This section addresses the challenges of labeling tasks with LMs, the various reasons behind these difficulties, and future research directions to lift these limitations.

\begin{itemize}[nosep]
    \item \textbf{Tasks challenging to label with LLMs}:  tasks which require subjective evaluation \cite{dillion2023can}, deep domain or culture-specific knowledge \cite{thirunavukarasu2023large,wang2024seaeval}.
    \item \textbf{Limitations of LLMs in labeling:} biases towards various social groups \cite{gallegos2024bias}; compute requirements \cite{kaplan2020scaling}; inevitable tendency to hallucinate inaccurate labels \cite{xu2024hallucination}; model collapse caused by training on synthetic data \cite{shumailov2024ai}.
\end{itemize}

\subsection*{Part 7: Hands-on session: Hybrid data annotation (30 min)}

\textbf{Overview}  In this hands-on session, we will implement a hybrid approach on a real-world dataset and demonstrate improvements in annotation quality.



%

\begin{itemize}[nosep]
    \item \textbf{Task}: implementing a hybrid approach on a real-world dataset to show how combining human labeling with LLM-generated labels can produce high-quality annotation.
    \item \textbf{Data}:  an open dataset for multiclass classification with high-quality expert annotations as ground truth.
    \item \textbf{Human labeling}: obtain noisy labels via Toloka's crowd platform. 
    \item \textbf{Labeling with LLMs}: (i) prompt LLMs in zero-shot and few-shot setups, and (ii) fine-tune a custom LLM on a held-out dataset portion for label generation.
    \item \textbf{Hybrid labeling}: demonstrating strategies from Part 5 to combine LLM and human labels, evaluating their performance wrt ground truth.
\end{itemize}

\paragraph{Setup} The demonstration will be conducted using Google Colab Notebooks. All materials will be published in advance for the audience to download and reproduce. To save time, the crowdsourcing annotation and the model training will be performed beforehand. The annotated dataset and annotation guidelines will be published for further uptake. Attendees can follow along without needing to code independently.

\section{Tutorial Information}

\noindent \textbf{Tutorial type and duration} This is a \textit{cutting-edge, half-day, in-person} tutorial. The increasing demand for labeled data justifies discussing automatic and hybrid approaches for data annotation. The widespread adoption of LLM-based technologies introduces novel data labeling approaches that accelerate annotation, reduce human workload in redundant tasks, and allow human workers to focus on complex instances. This tutorial differs from prior tutorials on data labeling \cite{suhr-etal-2021-crowdsourcing, drutsa-etal-2021-crowdsourcing}, which deal with labeling by humans only, and from prior tutorials on human-AI interaction \cite{yang-etal-2024-human,nips2023-tutorial}, which focus on a broad range of tasks in which AI and humans can interact.

\noindent  \textbf{Tutorial audience and pre-requisites} This tutorial is aimed at NLP practitioners who develop novel datasets and data collection protocols. While it provides an overview of the current best practices and state-of-the-art approaches in hybrid labeling with LLMs and human-in-the-loop, it is designed to fit an entry-level audience. The tutorial assumes a basic understanding of supervised classification, such as learning from labeled data and the concept of a classifier's confidence. Understanding of current large language models is recommended but not necessary.  We will introduce the necessary terms and notations in each section of the tutorial. The final part of the tutorial includes a high-level code walkthrough for those interested in implementing hybrid approaches on their own.

\noindent \textbf{Software} Throughout the tutorial, we will use the Python language and Google Colab with GPU access as an environment.

\noindent  \textbf{Approximate count} We expect this tutorial to attract over 100 attendees, reflecting the increasing demand for labeled data in both industrial and research settings.

\noindent  \textbf{Tutorial materials} Tutorial resources are  made available online for further uptake.

\section{Diversity Considerations}

The topic of the tutorial is linked to the ultimate goal of achieving a representation of diversity in NLP. We place special emphasis on the limitations of current automatic approaches to data labeling, particularly in tasks where the diversity of annotator representation is important. We discuss strategies to ensure ethical and fair use of manual labor in data annotation that guarantees better work conditions for annotators. Finally, we emphasize the importance of efficient approaches that reduce compute load.  

The presenters are diverse in terms of gender, language, career stages, background, occupation, affiliation, and location.

\section{Reading List}
\begin{enumerate}
    \item \textbf{Labeling with humans}: handbook  \cite{ide2017introduction}, quality control with CrowdKit \cite{ustalov2024learning}
    \item \textbf{Synthetic data and labeling with LLMs}: surveys \cite{honovich-etal-2023,tan2024large}, 
    \item \textbf{Active learning}: surveys \cite{schroder2020survey,zhang2022survey}, active learning with LLMs \cite{dor2020active,zhang2023llmaaa}
    \item \textbf{Hybrid labeling}: surveys  \cite{wang2021putting,wu2022survey}, blog entry \cite{toloka2024llms} 
\end{enumerate}

\section{Presenters}

\paragraph{\href{https://scholar.google.com/citations?hl=en&user=G0lCb3wAAAAJ}{Ekaterina Artemova}} is a Senior Research Scientist at Toloka AI, holding a PhD from FRC CSC RAS. Her research focuses on data-centric NLP, specifically benchmarking strategies, low-resource settings, and LLM evaluation. Ekaterina has published in leading NLP and AI conferences and journals, including NLP, LRE, EMNLP, and *ACL. She has co-organized the 1st NLP Power! workshop at ACL '22, a tutorial on artificial text detection at INLG '22, and multiple shared tasks at the Dialogue conference.

\paragraph{\href{https://scholar.google.com/citations?user=0_u3VUUAAAAJ&hl=en&oi=ao}{Akim Tsvigun}} is a Senior ML Architect at Nebius AI. He has a strong publication record in top-tier NLP conferences, including ACL, EMNLP, and NAACL. His research primarily focuses on uncertainty estimation and active learning in natural language processing tasks, with an emphasis on improving model robustness and efficiency.

\paragraph{\href{https://scholar.google.com/citations?view_op=list_works&hl=en&hl=en&user=7JjqFPoAAAAJ&sortby=pubdate}{Dominik Schlechtweg}} is an independent research group leader at the Institute for Natural Language Processing. He leads the Stuttgart group of the 6-year research program `Change is Key!' where his group is responsible for the automation of semantic annotation processes and the application of such processes to the task of detecting semantic changes. His group develops several online text annotation systems. He organized several shared tasks on semantic change detection including SemEval-2020 Task 1 and created various Word-in-Context datasets.

\paragraph{\href{https://scholar.google.com/citations?user=ifvqn8sAAAAJ&hl=en&oi=sra}{Natalia Fedorova}} is the Head of Academic Programs at Toloka. She is in charge of establishing long-term collaboration with universities, research labs, and educational organizations. Natalia co-organized a series of workshops and tutorials about crowdsourcing at NeurIPS, RecSys, ECIR, and WSDM conferences. She has a Master's degree in social psychology. Her research interests include the efficiency of education, the role of personality in education, role of humans in AI development.

\paragraph{\href{https://scholar.google.com/citations?hl=en&user=yLrJBmIAAAAJ}{Konstantin Chernyshev}} is a Research Scientist at Toloka AI, specializing in data quality and LLM evaluation. He led the development of the U-MATH dataset, a benchmark for university-level mathematics in AI. Konstantin holds a Master's degree in 'Language and Communication Technologies' from the University of Groningen and Saarland University and has published work in ACL. His research interests include explainable AI, quality-centric training, and model efficiency.

\paragraph{\href{https://scholar.google.com/citations?user=7o0HMXsAAAAJ&hl=en&oi=ao}{Sergei Tilga}} is the Head of R\&D at Toloka AI, with over 7 years of industrial experience applying ML models and AI systems in consumer products and production processes. Sergei obtained a Master's degree from MSU. His research interests encompass AI model alignment, LLM evaluation, and data quality.

\paragraph{Boris Obmoroshev} is the Data and R\&D Director at Toloka, with over 8 years of industrial experience in practical Data Science applications. He leads a team of Data Science and Machine Learning engineers, focusing primarily on adopting modern natural language processing technologies for large-scale data labeling projects. His main research interests are data quality and AI model evaluation. Boris obtained a Master's degree from MSU.

\section{Ethics Statement}
During the tutorial, we will address the limitations of using LLMs for data annotation, such as their predisposition to social biases and significant compute requirements. We will highlight the importance of respectful and responsible use of human annotators, ensuring fair compensation, informed consent, clear instructions, and reasonable working hours to avoid exploitation in any form.

While we acknowledge the risks of over-relying on automated approaches for data labeling, we believe the benefits of hybrid strategies outweigh these risks. This tutorial emphasizes leveraging a balanced mix of human and automated labeling techniques to ensure higher quality and fairer data production.

\paragraph{Use of AI-assistants} We improve and proofread the text of this paper using Grammarly\footnote{\href{https://app.grammarly.com}{\texttt{grammarly.com}}} to correct grammatical, spelling, and style errors and paraphrasing sentences. Therefore, specific segments of our proposal can be detected as AI-generated, AI-edited, or human-AI-generated.

\section{References}\label{sec:reference}

\bibliographystyle{lrec-coling2024-natbib}
\bibliography{custom}

\end{document}